\documentclass{article}
\PassOptionsToPackage{numbers, compress}{natbib}
\usepackage{float}

\usepackage[preprint]{neurips_2024}

\usepackage{xcolor}         
\usepackage{graphicx}
\usepackage{multirow}
\usepackage{caption}
\usepackage{amsmath}

\title{A 7K Parameter Model for Underwater Image Enhancement based on Transmission Map Prior}

\author{
  Fuheng Zhou $^1$  \And Dikai Wei$^1$ \And Ye Fan$^1$ \And Yulong Huang$^1$ \And Yonggang Zhang $^1$ \thanks{Corresponding author: Yonggang Zhang, Email:zhangyg@hrbeu.edu.cn}   \\
   1 College of Intelligent Systems Science and Engineering\\
  1 Harbin Engineering University\\
  1 Harbin 150001, China \\
  \texttt{zhangyg@hrbeu.edu.cn} \\}

\begin{document}
\maketitle
\begin{abstract}
Although deep learning based models for underwater image enhancement have achieved good performance, they face limitations in both lightweight and effectiveness, which prevents their deployment and application on resource-constrained platforms. Moreover, most existing deep learning based models use data compression to get high-level semantic information in latent space instead of using the original information. Therefore, they require decoder blocks to generate the details of the output. This requires additional computational cost. In this paper, a lightweight network named lightweight selective attention network (LSNet) based on the top-k selective attention and transmission maps mechanism is proposed. The proposed model achieves a PSNR of 97\% with only 7K parameters compared to a similar attention-based model. Extensive experiments show that the proposed LSNet achieves excellent performance in state-of-the-art models with significantly fewer parameters and computational resources. The code is available at \textcolor{blue}{https://github.com/FuhengZhou/LSNet.}
\end{abstract}

\section{Introduction}
\label{sec:intro}
Underwater images are often affected by light degradation, which leads to haze, color casts, or light absorption \cite{1-jian2021underwater}. These degradation factors result in lower image quality. These low quality underwater images affect underwater operations. Therefore, underwater image enhancement techniques can greatly assist these underwater operations. Underwater image enhancement methods can be divided into traditional methods, physical model methods, and deep learning methods \cite{2-shi2022integrating}. Traditional methods adjust the pixel of the image by stretching, sharpening, or straightening pixels \cite{3-hitam2013mixture, 4-ancuti2017color, 5-li2015underwater}. Physical model methods reverse the underwater image degradation by prior distribution information or establish models to simulate the degradation of light \cite{6-sathya2015underwater, 7-wen2013single, 8-drews2013transmission}. However, traditional methods cannot adjust their parameters to deal with different scenes. Although some physical model methods can enhance images by simulating light degradation, these methods could not simulate all degradation factors. Deep learning methods learn the mapping relationship between raw images and reference images \cite{9-perez2017deep}. These methods can automatically catch useful features which suitable for different scenarios, so they have become mainstream methods. 
\begin{figure}[tb]
\hspace{-0.3cm}
  \includegraphics[height=10.5cm]{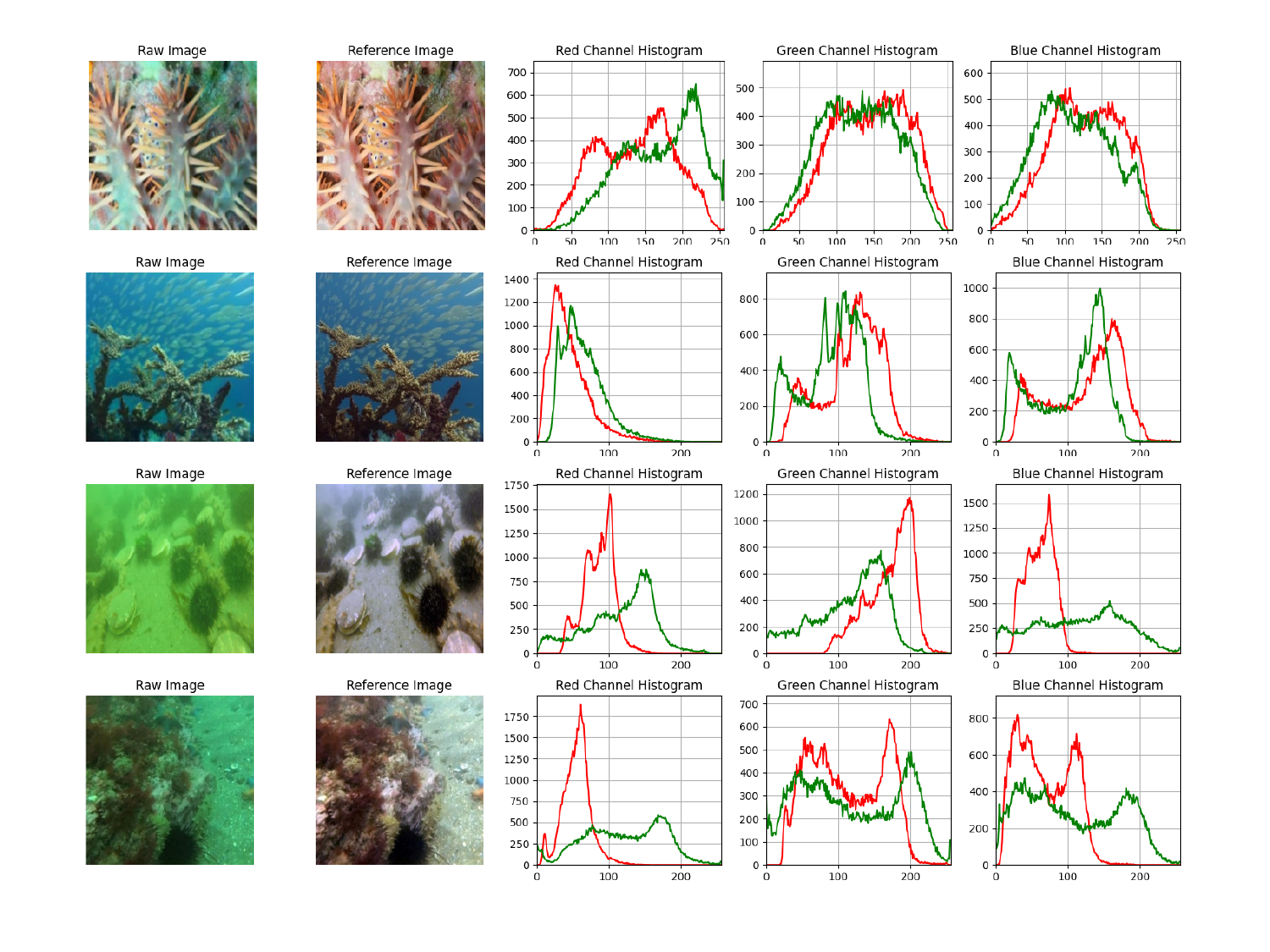}
  \caption{The visualization of raw and reference images and their histograms of red, green, and blue channels. In each histogram picture, the red curve represents the raw image and the green curve represents the reference image. Compared with reference images, it can be easily found that the pixels of the red channel and blue channel of the raw image are clustered on the left, while the green channel is more uniform.}
  \label{fig:figure1}
\end{figure}

In the underwater environment, the resource-constrained equipment facilitates are more suitable for underwater tasks than the large equipment. However, existing CNN-based models or transformer-based models require a large number of parameters. This limits the application of the deep learning model on resource-constrained platforms. One solution is to use lightweight models, such as UWCNN \cite{34-li2020underwater} and Shallow-uwnet \cite{20-naik2021shallow}. Unfortunately, these models are limited by the type of scene attenuation or few parameters, which results in lower accuracy. Because most of these models use an encoder-decoder structure to obtain high-level features in latent space and then decode the features to obtain the results or they do not use the original information directly and lack statistical and physical priors. Therefore, they use a lot of parameters for latent space feature extraction and reconstruction. These problems pose challenges for underwater image enhancement models on lightweight and effectiveness.

To address these challenges, inspired by transmission-maps-based underwater image restoration methods, LSNet based on the top-k selective attention and transmission maps mechanism is proposed. LSNet is an efficient underwater enhancement network that does not use latent space. LSNet decomposes clean underwater images (reference images) into raw images, estimated compensation images, and over exposure attenuation images. The contributions of this paper are as follows:

\begin{itemize}
\item A new underwater image enhancement model is proposed that uses transmission maps prior and deep learning to get the enhanced result without data compression in the latent space.
\item The lightweight attention mechanism based on top k selection is introduced, which can find regions with similar degradation.
\item The proposed model only uses 7K parameters to obtain results close to the SOTA model, which greatly reduces the number of parameters.

\end{itemize}
The rest of this paper is organized as follows: Section \ref{sec:related} presents the related works of underwater image enhancement models. Section \ref{sec:method} shows the details of the LSNet model. In section \ref{sec:exp}, the qualitative and quantitative experiments are introduced. Section \ref{sec:con} concludes this paper. 

\section{Related Work}
\label{sec:related}
\subsection{Deep Learning-based Underwater Image Enhancement Methods}
With the success of deep learning in various tasks, researchers have begun to use the deep learning method to the underwater image enhancement task. Wang et al. proposed a CNN-based end-to-end framework of underwater image enhancement for color correction and dehazing, named UIE-Net \cite{10-wang2017deep}. Anwar et al. used deep transmission maps to synthesize different degradation scenarios and proposed a set of lightweight networks to enhance the different degradation scenarios \cite{11-anwar2018deep}. Li et al. proposed an underwater image enhancement benchmark (UIEB), which includes 950 real underwater images. He also proposed a Water-Net which trained on this benchmark \cite{12-li2019underwater}. Guo et al. proposed a novel multi-scale dense generative adversarial network (GAN) to enhance underwater images \cite{13-guo2019underwater}. Islam et al. proposed a real-time underwater image enhancement model based on conditional generative adversarial networks \cite{14-islam2020fast}, and formulated an objective function to evaluate perceived image quality based on global content, color, local texture, and style information. They also proposed a large-scale dataset called EUVP. Li et al. proposed a multi-color space encoder network that enriches the diversity of feature representations by merging features from different color spaces into a unified structure \cite{15-li2021underwater}. Fu et al. proposed a novel probabilistic network that combines conditional variational auto-encoders with adaptive instance normalization to construct augmented distributions \cite{16-fu2022uncertainty}. Peng et al. proposed a U-shape transformer network and designed a new loss function that combines multiple color spaces \cite{17-peng2023u}. Although these networks achieve good results, they require numerous parameters and a high amount of computation, making them impractical to deploy in resource-constrained devices.

\subsection{Lightweight Underwater Image Enhancement Methods}
For underwater environments, lightweight underwater image enhancement networks are easier to use and deploy on resource-limited platforms and therefore have become a hot research topic recently. Yang et al. proposed a lightweight adaptive feature fusion network called LAFFNet \cite{18-yang2021laffnet}. The model is an encoder-decoder model with multiple adaptive feature fusion modules. Lyu et al. proposed a simple and effective two-stage image enhancement network \cite{19-lyu2022efficient}. Naik et al. proposed a shallow neural network architecture, named Shallow-uwnet \cite{20-naik2021shallow}. It maintains performance and has fewer parameters. Guo et al. proposed a simple normalization-based u-shape UIE network, named NU2Net \cite{22-guo2023underwater}. The network used URanker for auxiliary training. Jiang et al. proposed an underwater image enhancement method based on a fast fourier convolutional network \cite{23-jiang2023five}, which keeps the network parameters very low. Although these networks have achieved some good performance, they still have computational redundancy because they do not share parameters in similar attenuation areas and use reconstruction modules for detailed recovery. Therefore, a more lightweight model still requires further research.

\begin{figure}[tb]
\hspace{-0.3cm}
\centering
  \includegraphics[height=7.5cm]{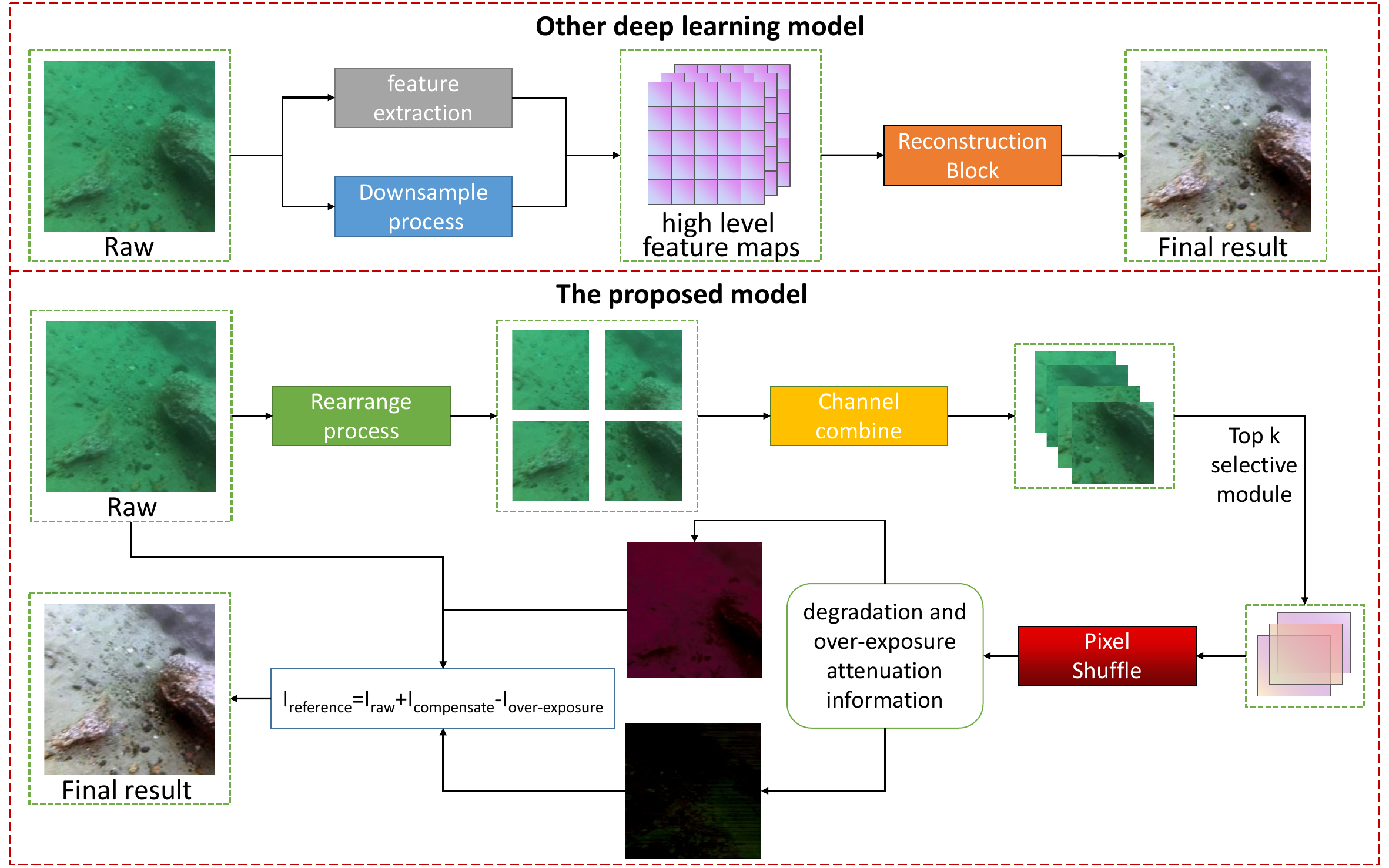}
  \caption{The comparison of other deep learning methods and LSNet. It can be found that LSNet does not have any downsampling, so the reconstruction module for high-level features is not needed, while other methods need to generate lost details due to the loss of information caused by the downsampling process.}
  \label{fig:figure2}
\end{figure}
\section{Methodology}
\label{sec:method}
This chapter introduces the details of LSNet. First, the inspiration and idea of LSNet are introduced. Secondly, the details of the model structure of LSNet are introduced. Finally, the improved top-k attention mechanism module is introduced.
\subsection{Inspiration of LSNet}
\label{sec:design}
As mentioned in \ref{sec:intro}, the idea of LSNet comes from transmission-maps theory. The transmission maps can be used as a method for underwater image enhancement because such methods assume that areas with the same transmission rate have similar degradation, such as similar brightness and color degradation. The degradation of underwater images from a medium can be described as the interactions between the medium, the environment (scene), and light \cite{8-drews2013transmission}. Jaffe-McGlamery model described a function to describe these interactions by three factors: direct illumination ($I_d$), forward-scattering ($I_{fs}$), and backscatter ($I_{bs}$) \cite{37-jaffe1990computer,38-mcglamery1980computer}, as shown in equation \ref{eq:e1}:
\begin{equation}
\centering
I_T = I_d + I_{fs} + I_{bs}
\label{eq:e1}
\end{equation}
Where $I_T$ is the image captured by the camera (raw image). The forward scatter is the light reflected by the object. It will disappear by the scatter and absorption. Due to \cite{39-schechner2005recovery}, the backscatter is the main reason for image contrast degradation, so most methods neglect the forward-scattering. The direct illumination can be defined as \ref{eq:e2}:
\begin{equation}
\centering
I_d = Je^{-\eta d} = Jt
\label{eq:e2}
\end{equation}
Where $J$ and $d$ are the scene radiance and the distance, $\eta$ is the attenuation coefficient, $t$ is the medium transmission. In natural scenes, $\eta$ consists only of scattering. In the underwater environment, $\eta$  consists of scattering and absorption \cite{37-jaffe1990computer}. The backscatter comes from the interaction between the illumination sources and particles in the medium, it can be described as \ref{eq:e3}:
\begin{equation}
\centering
I_{bs} = A(1-e^{-\eta d}) = A(1-t)
\label{eq:e3}
\end{equation}
So, the degradation of underwater images from a medium can be described as \ref{eq:e4}:
\begin{equation}
\centering
I(x) = J(x)e^{-\eta d}(x)+A(1-e^{-\eta d}(x))
\label{eq:e4}
\end{equation}
Where $I(x)$ is the raw degradation image.
In the DCP \cite{30-he2010single}, He et al. use the dark channel for calculating the transmission maps, as shown in \ref{eq:e5}:
\begin{equation}
\begin{split}
&J^{dark}(x) = \min_{y\in \Omega(x)} (\min_{c\in \Omega(R,G,B)}J^c(y)) \\
&\tilde{t} = 1 -\min_{y\in \Omega(x)} (\min_{c\in \Omega(R,G,B)}\frac{I(y)}{A} )
\end{split}
\label{eq:e5}
\end{equation}
They assume that the intensity of $J$ is low and tends to zero ($J^{dark} \to$ 0) in a clear image except for the sky region. Then, they use \ref{eq:e6} to get the final clear image. 
\begin{equation}
\begin{split}
J(x) = \frac{I(x)-A}{max(t(x),t_0)}+A
\end{split}
\label{eq:e6}
\end{equation}
However, estimating atmospheric illumination $A$ and $t$ is very difficult because they use simplified models and the forward scattering is ignored (For example, they use the top 0.1\% of the pixels in the image as the value of A, and assume that the local area has the same transmittance). But they still provide two things: (1) A clean image can be obtained by the degraded image (raw image) and the transmission map; (2). Locations with the same transmission value have the same decay (the value of t(x) is the same). Therefore, according to the Fig. \ref{fig:figure1} and the equation \ref{eq:e1}, the attention is used for estimating the region-related features, the equation \ref{eq:e4} is changed to the following equation:
\begin{equation}
\begin{split}
J(x) = I(x) + \{I(x)e^{\eta d} + A\} - \{I(x) + Ae^{\eta d} + I_{fs}e^{\eta d}\}
\end{split}
\label{eq:e7}
\end{equation}
In Fig.\ref{fig:figure1}, a phenomenon was discovered that compared with the reference image, the red channel and blue channel in most underwater raw images have information lost. It can be seen that in the histogram, there are more pixels with lower values in the red channel and blue channel than the green channel. In the green channel, the image has a green color cast and there are more pixels with higher values in the green channel than the red and blue channels, which corresponds to equation \ref{eq:e7}. Where the $I(x)e^{\eta d} + A$ is used as compensation maps, the $I(x) + Ae^{\eta d}$ is used as over exposure maps. Thus, the clear image is decomposed into the raw image, the compensated image, and the over exposure attenuation image, as shown in equation \ref{eq:eq1}. 
\begin{equation}
\centering
J(x) = I(x) + I_{compensate\_image}(x) - I_{exposed\_image}(x)
\label{eq:eq1}
\end{equation}
This is the basic theory of the LSNet and learning these two types of images can reduce the number of parameters in the model.

\subsection{Model Structure}
\begin{figure}[tb]
\hspace{-0.3cm}
\centering
  \includegraphics[height=7.5cm]{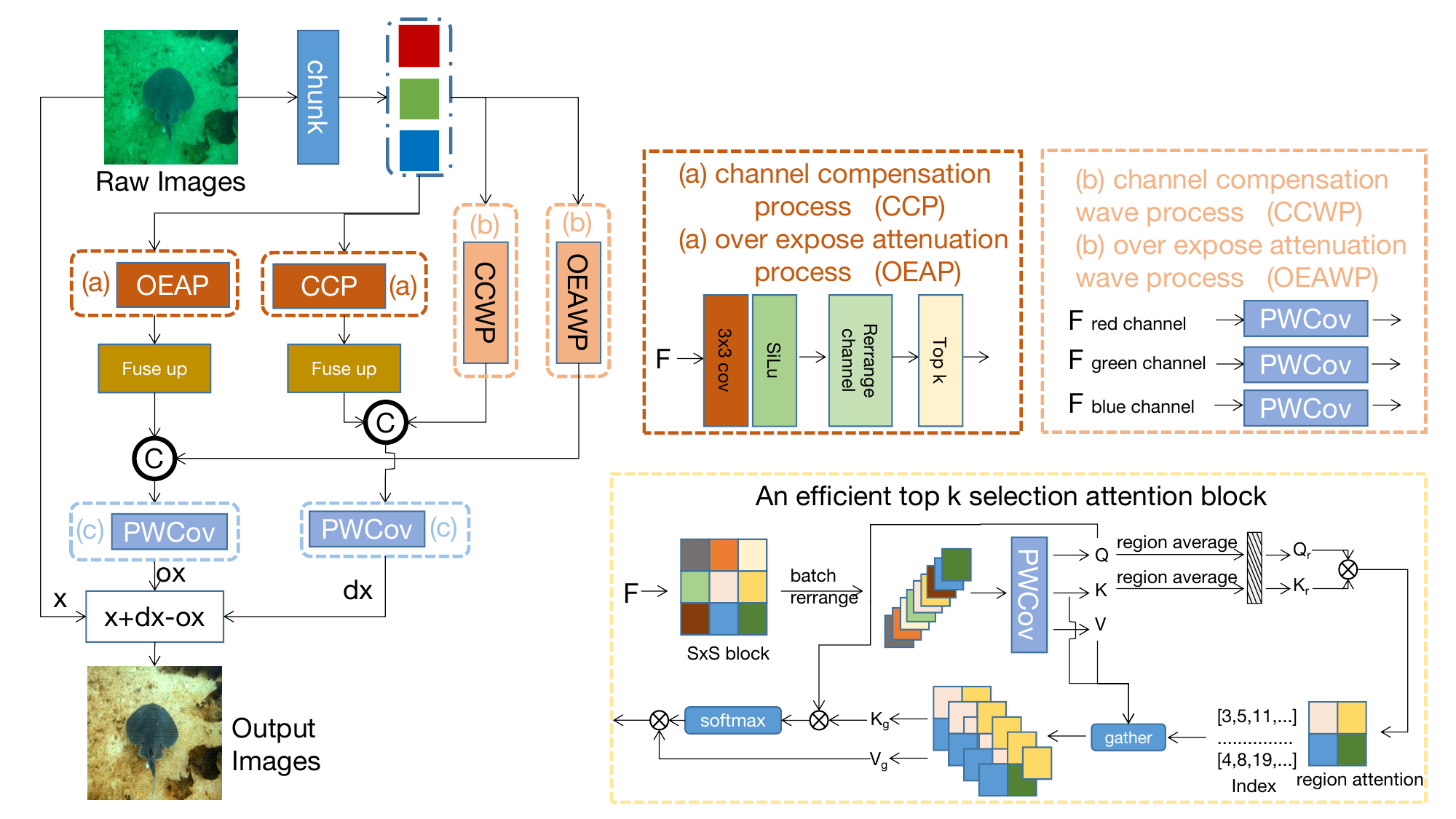}
  \caption{The detail of the LSNet structure. The LSNet uses the chunk function to chunk the inputs from the channel to batch to use the GPU more efficiently.}
  \label{fig:structure}
\end{figure}

LSNet is a lightweight model, so there are not many channels in the data flow process. To keep the model lightweight, all convolution kernels use pointwise convolution (PW conv) except for the 3x3 convolution kernel that provides position information. The data flow process of the LSNet is as follows: First, the raw image is chucked into red, blue, and green channels. The purpose of this is because the red, blue, and green channels have different attenuation or over-exposure coefficients due to different wavelengths. Therefore, processing each color channel individually could capture the wavelength attenuation or exposure coefficients better. LSNet transfers them to 16 channels respectively to obtain more nonlinear features. This process loses the attenuation and over exposure correlation between the three channels. Therefore, the 3x3 convolution kernel replaces the position encoding and provides interactivity between the three channels. Afterward, the obtained three-channel feature maps of attenuation or exposition are split along the channel dimension, as shown in Fig. \ref{fig:figure2}. After that, the improved top-k attention mechanism is used to find the correlation between regions, and the obtained features are concatenated with the previous 16 channels' nonlinear features information by using sub-convolution, and then using pointwise convolution to produce the channel compensation image and over-exposition attenuation image. Finally, the final enhanced image is obtained by the original image information, compensation images, and over exposure attenuation images, as shown in Fig \ref{fig:structure}. 

\subsection{Improvement Top-k Selection Attention Module}
To identify similar degraded regions (similar to the transmission map value), an improved attention module based on the top-k selection mechanism is proposed \cite{24-zhu2023biformer}. First, a 3x3 convolution kernel with 3 input and output channels is used instead of the learnable position encoding for labeling spatial locations, as shown in equation \ref{eq:e8}:
\begin{equation}
\centering
E_{PE} = Chunk\{\delta (Cov_{3x3}I(x))\}
\label{eq:e8}
\end{equation}
Secondly, the input feature map is divided into four blocks along the batch dimension. The aim of this is to use GPU parallel computing so that different parts of the same image become different batches. They share and reduce the learnable parameters, as shown in \ref{eq:e9}. 
\begin{equation}
\centering
E_{to\_batch} = arrange(E_{PE})
\label{eq:e9}
\end{equation}
Pointwise convolution and batch normalization layers are then used instead of Linear modules to generate region queries, keys, and values. This is the only block in the entire model that uses batch normalization because this paper believes that batch normalization is good at identifying commonalities in features since regions with the same degree of degradation should have the similar QKV value, as shown in \ref{eq:e10}. 
\begin{equation}
\begin{split}
&Q, K, V = \delta\{BN(PW_{Cov}(E_{to\_batch})\}\\
&Q_{r}, K_{r} = Q_{s}.mean, K_{s}.mean\\
&A_{r}=Q_{r} * K_{r}, Index = topk(A_{r}, k)
\label{eq:e10}
\end{split}
\end{equation}
Where $BN$ is the batch normal, $PW_{Cov}$ is the Pointwise convolution, $E_{to\_batch}$ is that chunk the inputs from the channel to batch, $Q_{r}, K_{r}, A_{r}$ are the region of the Query, Key, and attention according to \cite{24-zhu2023biformer}. After the regional attention is generated, the most K-relevant regions are selected from all regions. Then fine-grained attention is calculated for the regions in each Q and the KV regions, as shown in equation \ref{eq:e11}. 
\begin{equation}
\begin{split}
&K_{topk} = gather(K, I_r)\\
&V_{topk} = gather(V, I_r)\\
&A = Softmax(Q * K_{topk}) * V_{topk}
\end{split}
\label{eq:e11}
\end{equation}
Where $K_{topk}, V_{topk}$ are the fine-grained value of the region K and V. To keep it lightweight, MLP was removed since it requires significant computational resources. The entire process calculates the compensation image and the over exposure attenuation image in parallel, as shown in Fig. \ref{fig:structure}.
\section{Experiment}
\label{sec:exp}
In this section, the experimental details are first introduced, which includes the parameter settings and dataset selection. Then, the evaluation metrics are introduced. Next, the comparative models and results for this experiment are introduced. Finally, the ablation experiments and analysis are introduced.
\subsection{Experimental Details}
LSNet is implemented by using PyTorch and trained on an NVIDIA RTX 3090 (24GB). During training, the model is trained for 1000 epochs. The Adam optimizer and L1 loss are used for training the proposed model. For testing datasets, 515 paired test images from the EUVP dataset are selected \cite{14-islam2020fast}, named Test-E. For the LSUI dataset \cite{17-peng2023u}, following the authors' setting, 3,879 images are randomly selected as the training dataset, and the remaining 400 images are used as the testset, named Test-L. In the RUIE dataset \cite{40-liu2020real}, non-paired data of green and blue types are used as the test dataset, named Test-R. The U45 dataset is also used for testing, named Test-U \cite{41-li2019fusion}.

\begin{figure}[tb]
\centering
  \includegraphics[height=8.5cm]{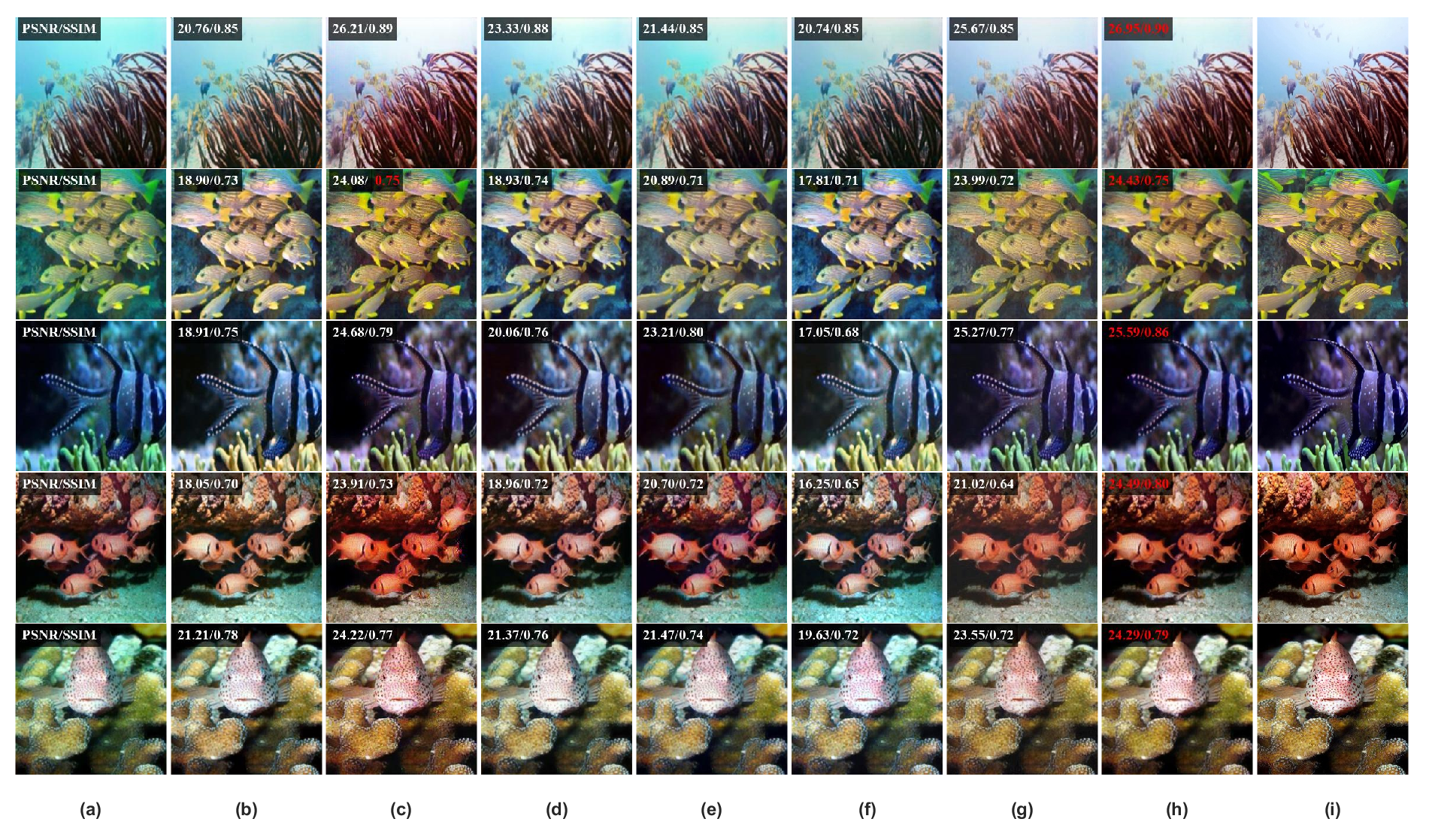}
  \setlength{\abovecaptionskip}{-0.5cm}
    \setlength{\belowcaptionskip}{-0.3cm}
  \caption{The visualization results of the LSNet. From left to right: (a) origin, (b) FA+Net, (c) FUnIE, (d) NU2Net, (e) P2CNet, (f) UIEC2Net, (g) U-shape, (h) LSNet, (i) The ground truth. It can be seen that LSNet achieves good results while maintaining extremely low parameters.}
  \label{fig:figure3}
\end{figure}

\subsection{Evaluation Metrics}
To comprehensively evaluate the results produced by each model, the Peak Signal-to-Noise Ratio (PSNR), the Structural Similarity Index (SSIM) \cite{26-wang2004image}, the Underwater Color Image Quality Evaluation (UCIQE) \cite{27-yang2015underwater}, the Underwater Image Quality Metric (UIQM) \cite{28-panetta2015human}, the underwater image colorfulness measure (UICM) \cite{28-panetta2015human}, the underwater image sharpness measure (UISM) \cite{28-panetta2015human}, and the natural image quality evaluator (NIQE) are used as performance metrics for image quality \cite{25-mittal2012making}. The PSNR is a full-reference image quality evaluation metric based on errors between corresponding pixels. A higher PSNR score indicates a similar output. The SSIM measures the visual quality of three features of an image: brightness, contrast, and structure. A higher SSIM value indicates a higher similarity between the enhanced and reference images. The UCIQE mainly measures the detail and color recovery of distorted images. The UIQM is used to evaluate color, sharpness, and contrast. The UICM and the UISM are part of the UIQM. The NIQE is based on measurable deviations from statistical regularities observed in natural images. However, it is worth noting that deep learning models are affected by the training dataset. It is difficult to accurately evaluate the performance of the model under subjective evaluation methods. Therefore, this paper focuses on the results under the PSNR and the SSIM evaluation functions.

\subsection{Experiment results}

\begin{figure}[htbp]
  \begin{minipage}[t]{0.45\textwidth}
  \vspace{1cm} %
    \resizebox{\textwidth}{!}{%
    \begin{tabular}{c|llll}
\hline
\multicolumn{1}{l|}{} & \multicolumn{1}{c}{flops $\downarrow$} & \multicolumn{1}{c}{param $\downarrow$} & \multicolumn{1}{c}{time $\downarrow$} &  \\ \hline
SMBL \cite{29-wang2019experimental}             & \multicolumn{1}{c}{/}     & \multicolumn{1}{c}{/}     & 1.3561s                  &  \\
DCP \cite{30-he2010single}                   & \multicolumn{1}{c}{/}     & \multicolumn{1}{c}{/}     & 1.3091s                  &  \\
FA+Net \cite{23-jiang2023five}                & {\color[HTML]{0000FF}585.2378M}                 & {\color[HTML]{0000FF} 0.009M}                   & 0.0084s                  &  \\
FUnIE \cite{14-islam2020fast}                 & 10.2388G                  & 7.0196M                   & {\color[HTML]{FF0000}0.0011s}                  &  \\
P2CNet \cite{31-rao2023deep}                & 5.1796G                   & 2.0676M                   & 0.0086s                  &  \\
PUIE \cite{16-fu2022uncertainty}                  & 30.0939G                  & 1.4010M                   & 0.1484s                  &  \\
Water-Net \cite{12-li2019underwater}             & 142.9039G                 & 1.0907M                   & 0.1297s                  &  \\
Ucolor \cite{15-li2021underwater}                & 2.8053T                   & 148.7712M                 & 2.5584s                  &  \\
U-shape \cite{17-peng2023u}               & 2.9835G                   & 22.8172M                  & 0.0951s                  &  \\
UWCNN \cite{34-li2020underwater}                 & 5.2299G                   & 0.04M                  & {\color[HTML]{0000FF} 0.0068s}                  &  \\
UWGAN \cite{35-wang2019uwgan}                 & {\color[HTML]{FF0000}3.8481M}                   & 1.9257M                   & 0.0717s                  &  \\
UIEC2Net \cite{36-wang2021uiec}              & 24.3673G                  & 2.0400M                   & 0.0515s                  &  \\
NU2Net \cite{22-guo2023underwater}                & 10.486G                   & 3.146M                    & 0.0425s                  &  \\
{\color[HTML]{FF0000}LSNet(Ours)}               &  712.490M                  & {\color[HTML]{FF0000}0.007M}                    & 0.0075s                  &  \\ \hline
\end{tabular}
}
  \end{minipage}%
  \begin{minipage}[t]{0.55\textwidth}
    \vspace{-0.5cm} %
    \hspace{-1.5cm} %
    \rotatebox{-90}{\includegraphics[width=\textwidth]{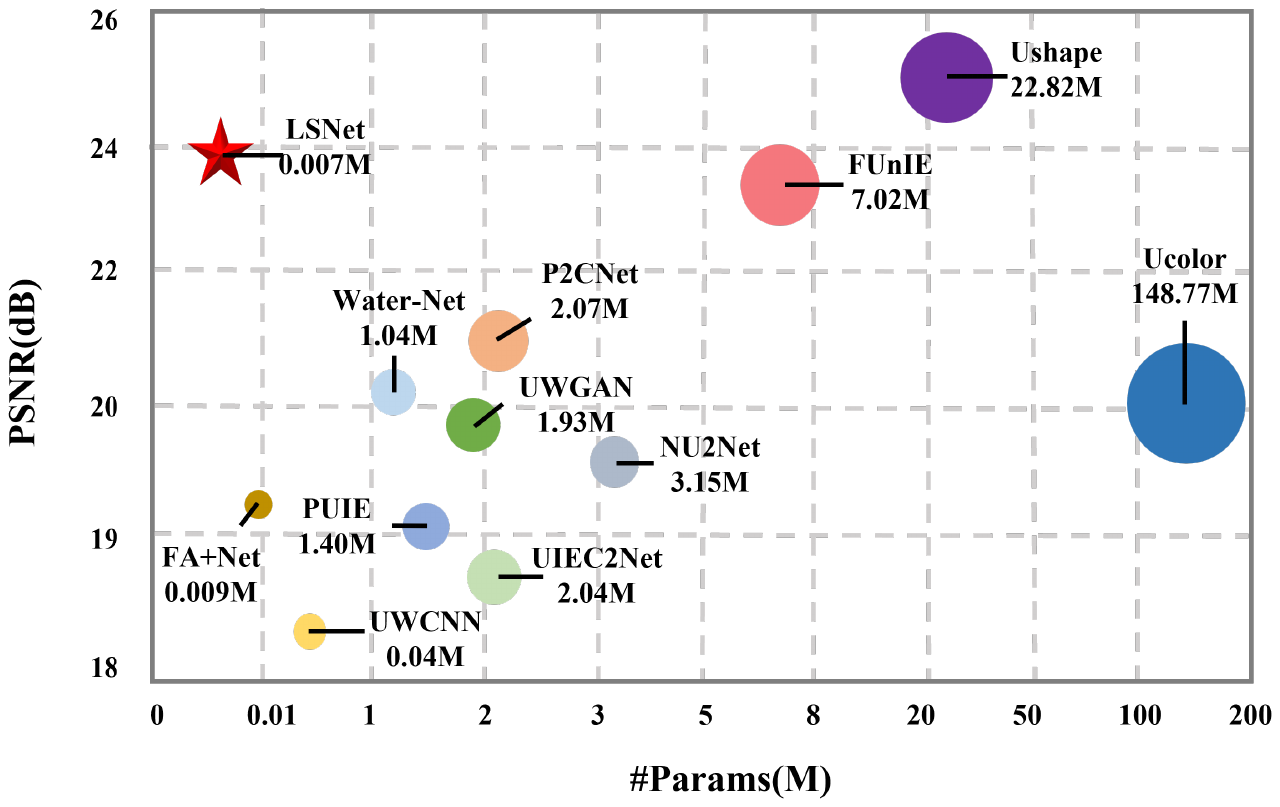}}
    \label{fig:efficientfig}
  \end{minipage}%
    \setlength{\abovecaptionskip}{-1.3cm}
    \setlength{\belowcaptionskip}{-0.3cm}
  \caption{The result of the LSNet and the other different models. The left table represents the results of flops, param, and time of the models. The right part represents the PSNR and parameters of the models. The proposed model LSNet achieves competitive results compared to the similar attention-based model U-shape (TIP 23) by only 0.03\% parameters.}
  \label{tab:effcient}
\end{figure}
To verify the LSNet with various models, SMBL \cite{29-wang2019experimental}, CLAHE \cite{3-hitam2013mixture}, DCP \cite{30-he2010single}, FA+Net \cite{23-jiang2023five}, FUnIE \cite{14-islam2020fast}, P2CNet \cite{31-rao2023deep}, PUIE \cite{16-fu2022uncertainty}, Water-Net \cite{12-li2019underwater}, Ucolor \cite{15-li2021underwater}, U-shape \cite{17-peng2023u}, UWCNN \cite{34-li2020underwater}, UWGAN \cite{35-wang2019uwgan}, UIEC2Net \cite{36-wang2021uiec}, NU2Net \cite{22-guo2023underwater} and the proposed LSNet were selected. It can be seen that LSNet surpasses many state-of-the-art models by PSNR and SSIM. 

For the accuracy performance of the LSNet, because LSNet uses the same training dataset as the U-shape, the U-shape can be used as the baseline of LSNet. As shown in table \ref{tab:table1}, it can be seen that LSNet achieves competitive results to the similar attention-based model U-shape \cite{17-peng2023u} (TIP23) by only 0.03\% of its parameters numbers. In table \ref{tab:table1}, six scores of LSNet on Test-E are higher than U-shape (SSIM, PCQI, UIQM, UICM, UISM, and UICQE). The LSNet also has 4 higher scores in UIQM, UICM, UISM, and UICQE to the U-shape on Test-L. For the non-reference datasets, the LSNet also achieved competitive results. It should be emphasized that LSNet only uses 0.03\% of the U-shape parameters. For the efficient result experiments in Fig.\ref{tab:effcient}, it can be seen that LSNet has a lower number of parameters and running time compared to the second low parameter model FA+Net \cite{23-jiang2023five}. The LSNet has higher scores in PSNR, SSIM, PCQI, UIQM, and UISM on Test-E and higher scores in PSNR, UIQM, and UISM on Test-L. For the non-reference datasets, the LSNet has a higher score in UISM on Test-U, a higher score in UIQM, UISM on Test-R (UCCS\_blue), and a higher score in UIQM, UICM, UISM, UICQE and NIQE on Test-R (UCCS\_green). These results demonstrated that the LSNet has excellent performance. Some of the visualizations are shown in Fig. \ref{fig:figure3}.

\begin{table}[]
\caption{The results of EUVP and LSUI test dataset. For each column, the best score is heighted in red, the second is heighted in blue.}
\resizebox{\textwidth}{!}{%
\begin{tabular}{c|cccccccc|cccccccc}
\hline
 &\multicolumn{8}{c|}{Test-E} &\multicolumn{8}{c}{Test-L} \\ \cline{2-17} 
\multirow{-2}{*}{} &PSNR  &SSIM &PCQI &UIQM &UICM &UISM &UICQE &NIQE &PSNR &SSIM &PCQI &UIQM &UICM &UISM &UICQE &NIQE \\ \hline
SMBL \cite{29-wang2019experimental} &16.0219 &{\color[HTML]{333333} 0.6831} &0.7825 &{\color[HTML]{333333} 2.0606} &{\color[HTML]{FF0000} 10.5778} &{\color[HTML]{333333} 4.6861} &{\color[HTML]{FF0000} 0.6529} &9.7045 &17.0125 &0.7443 &0.8035 &{\color[HTML]{000000} 2.3963} &{\color[HTML]{FF0000} 9.2618} &{\color[HTML]{000000} 5.3666} &{\color[HTML]{FF0000} 0.6257} &8.9949 \\
CLAHE \cite{3-hitam2013mixture} &17.3374 &{\color[HTML]{333333} 0.7281} &{\color[HTML]{FF0000} 0.8392} &{\color[HTML]{333333} 2.8357} &{\color[HTML]{0000FF} 7.1165} &{\color[HTML]{FF0000} 6.9323} &0.6062 &9.5795 &17.0511 &{\color[HTML]{000000} 0.7813} &{\color[HTML]{FF0000} 0.8621} &2.9204 &{\color[HTML]{000000} 5.9593} &{\color[HTML]{FF0000} 6.9468} &0.5845 &8.9489 \\
DCP \cite{30-he2010single} &17.5568 &0.6978 &0.6469 &1.8916 &6.1321 &4.9905 &0.5885 &9.8687 &15.7382 &0.6916 &0.6262 &1.9228 &5.0947 &5.1292 &0.5655 &8.8287 \\
FA+Net \cite{23-jiang2023five} &19.1937 &0.7825 &0.7467 &2.8931 &6.5192 &6.4872 &0.6100 &9.0186 &21.2887 &0.8390 &0.7878 &2.9370 &5.7722 &6.5278 &0.5997 &8.3365 \\
{\color[HTML]{000000} FUnIE \cite{14-islam2020fast}} &{\color[HTML]{000000} 23.4088} &{\color[HTML]{000000} 0.8033} &{\color[HTML]{000000} 0.7053} &{\color[HTML]{000000} 2.8410} &{\color[HTML]{000000} 6.4075} &{\color[HTML]{000000} 6.6156} &{\color[HTML]{000000} 0.5932} &{\color[HTML]{FF0000} 7.3083} &{\color[HTML]{000000} 21.7114} &{\color[HTML]{000000} 0.7992} &{\color[HTML]{000000} 0.6665} &{\color[HTML]{000000} 2.8856} &{\color[HTML]{000000} 5.6562} &{\color[HTML]{000000} 6.6427} &{\color[HTML]{000000} 0.5749} &{\color[HTML]{FF0000} 7.3450} \\
P2CNet \cite{31-rao2023deep} &20.8831 &0.7873 &0.6352 &2.9345 &6.2360 &6.0301 &0.5855 &8.9342 &20.0803 &0.7815 &0.5777 &2.8616 &5.6620 &5.6816 &0.5744 &8.2385 \\
PUIE \cite{16-fu2022uncertainty} &19.0737 &0.7890 &{\color[HTML]{333333} 0.7393} &{\color[HTML]{0000FF} 3.0280} &5.8055 &6.5076 &0.5992 &10.0734 &20.9709 &{\color[HTML]{0000FF}0.8551} &{\color[HTML]{000000} 0.7966} &{\color[HTML]{0000FF} 3.0493} &5.1897 &6.5147 &0.5881 &9.0181 \\
Water-Net \cite{12-li2019underwater} &20.3262 &0.7826 &0.7012 &2.9011 &6.3472 &6.5561 &0.6040 &9.0653 &21.2815 &0.8305 &0.7475 &{\color[HTML]{000000} 2.9602} &{\color[HTML]{0000FF} 6.2749} &6.6613 &{\color[HTML]{0000FF} 0.6027} &8.6871 \\
Ucolor \cite{15-li2021underwater} &20.0741 &0.7919 &0.6881 &2.9877 &5.9864 &6.3873 &0.5905 &9.0068 &21.3310 &0.8405 &0.7101 &2.9916 &5.2818 &6.3360 &0.5779 &8.3824 \\
UWCNN \cite{34-li2020underwater} &18.3517 &0.7510 &0.5986 &2.8456 &5.6083 &5.9645 &0.5769 &9.0444 &17.9188 &0.7578 &0.5604 &2.7674 &4.1187 &5.6693 &0.5517 &8.3005 \\
UWGAN \cite{35-wang2019uwgan} &19.7931 &0.7806 &0.6365 &2.8268 &4.4013 &6.0946 &0.5783 &8.9799 &18.9527 &0.8033 &0.6253 &2.7573 &3.2697 &5.9741 &0.5596 &8.2635 \\
UIEC2Net \cite{36-wang2021uiec} &18.7946 &0.7712 &0.7361 &2.9824 &6.5193 &{\color[HTML]{333333} 6.5635} &{\color[HTML]{0000FF} 0.6101} &8.8727 &21.4047 &0.8337 &0.7866 &2.9870 &5.8449 &{\color[HTML]{000000} 6.5951} &{\color[HTML]{0000FF} 0.6027} &8.1522 \\
NU2Net \cite{22-guo2023underwater} &19.5696 &0.7913 &{\color[HTML]{333333} 0.7557} &{\color[HTML]{FF0000} 3.0453} &{\color[HTML]{333333} 6.4937} &{\color[HTML]{0000FF} 6.7554} &0.6024 &10.2399 &21.7613 &{\color[HTML]{000000} 0.8504} &{\color[HTML]{0000FF} 0.8189} &{\color[HTML]{FF0000} 3.0844} &{\color[HTML]{000000} 5.8160} &{\color[HTML]{0000FF} 6.8028} &0.5925 &9.0930 \\
{\color[HTML]{000000} U-shape \cite{17-peng2023u}} &{\color[HTML]{FF0000} 25.1402} &{\color[HTML]{0000FF} 0.8201} &0.6741 &2.9234 &5.8190 &6.3393 &{\color[HTML]{333333} 0.5783} &{\color[HTML]{0000FF} 7.7870} &{\color[HTML]{FF0000} 25.0272} &{\color[HTML]{FF0000} 0.8570} &0.7044 &2.9566 &5.1662 &6.3621 &{\color[HTML]{000000} 0.5750} &{\color[HTML]{0000FF} 7.7720} \\
{\color[HTML]{000000} LSNet} &{\color[HTML]{0000FF} 24.2323} &{\color[HTML]{FF0000} 0.8279} &{\color[HTML]{0000FF} 0.8251} &2.9683 &5.9487 &6.5916  &0.5856 &{\color[HTML]{333333} 10.3011} &{\color[HTML]{0000FF} 23.1611} & 0.8124 &0.6936 & 3.0300 & 5.2329 & 6.7320 & 0.5777 &16.3780 \\ \hline
\end{tabular}
}
\label{tab:table1}
\end{table}

\begin{table}[]
\caption{The results of U-45 and UCCS test dataset. For each column, the best score is heighted in red, the second is heighted in blue.}
\resizebox{\textwidth}{!}{%
\begin{tabular}{l|lllll|lllll|lllll}
\hline
 &\multicolumn{5}{c|}{Test-U} &\multicolumn{5}{c|}{Test-R (UCSS\_blue)} &\multicolumn{5}{c}{Test-R (UCSS\_green)} \\ \cline{2-16} 
\multirow{-2}{*}{} &UIQM &UICM &UISM &UICQE &NIQE &UIQM &UICM &UISM &UICQE &NIQE &UIQM &UICM &UISM &UICQE &NIQE \\ \hline
SMBL \cite{29-wang2019experimental} &2.5771 &{\color[HTML]{FF0000} 7.9800} &5.8188 &0.5979 &6.4889 &2.6104 &{\color[HTML]{FF0000} 8.3059} &5.9725 &{\color[HTML]{FF0000} 0.6082} &10.2206 &3.0418 &{\color[HTML]{0000FF} 3.7129} &6.0883 &0.5274 &7.3448 \\
CLAHE \cite{3-hitam2013mixture} &2.8316 &5.3072 &6.7864 &0.5710 &6.5759 &2.9866 &3.7043 &{\color[HTML]{FF0000} 7.1422} &0.5426 &9.7088 &2.8996 &3.0939 &6.0837 &0.5179 &7.4180 \\
DCP \cite{30-he2010single} &2.0718 &5.2360 &5.5565 &0.5635 &{\color[HTML]{0000FF} 5.9700} &1.9232 &1.7794 &5.5281 &0.4950 &8.4467 &1.4434 &1.8364 &4.0042 &0.4960 &7.8618 \\
FA+Net \cite{23-jiang2023five} &3.1950 &5.5823 &6.9986 &{\color[HTML]{0000FF} 0.5999} &6.6125 &3.1887 &4.0312 &6.9941 &0.5691 &9.0012 &2.9003 &2.1717 &5.8694 &0.5194 &{\color[HTML]{0000FF} 5.9200} \\
FUnIE \cite{14-islam2020fast} &2.9188 &4.8049 &6.9038 &0.5594 &8.0421 &3.1933 &3.6631 &6.8711 &0.5158 &8.9536 &2.9202 &3.5822 &6.3029 &0.5187 &7.4790 \\
P2CNet \cite{31-rao2023deep}&3.0367 &5.8167 &6.0241 &0.5638 &6.2720 &3.0000 &3.3127 &6.0142 &0.5062 &{\color[HTML]{0000FF} 7.6225} &2.2421 &2.5249 &3.6797 &0.5433 &7.8024 \\
PUIE \cite{16-fu2022uncertainty} &{\color[HTML]{0000FF} 3.2135} &4.7753 &6.8271 &0.5755 &6.5438 &3.2203 &3.4213 &6.8778 &0.5424 &9.1084 &2.9531 &2.4792 &6.0034 &0.5319 &7.4016 \\
Water-Net \cite{12-li2019underwater} &3.1259 &5.1291 &6.8928 &0.5850 &8.6052 &3.2014 &{\color[HTML]{0000FF} 4.6902} & 7.0143 &{\color[HTML]{0000FF} 0.5693} &10.2043 &{\color[HTML]{FF0000} 3.1912} &{\color[HTML]{FF0000} 5.6822} &{\color[HTML]{FF0000} 6.7437} &{\color[HTML]{FF0000} 0.5961} &8.5615 \\
Ucolor \cite{15-li2021underwater} &3.0746 &4.9112 &6.7030 &0.5749 &6.2625 &3.1206 &2.9511 &6.7757 &0.5236 &8.8055 &2.9820 &2.6396 &5.9267 &0.5279 &6.6051 \\
UWCNN \cite{34-li2020underwater} &2.8008 &2.9168 &5.7777 &0.5247 &6.0965 &3.0171 &2.4727 &6.1825 &0.4884 &{\color[HTML]{FF0000} 7.4558} &2.4011 &0.5333 &3.9404 &0.4732 &7.5138 \\
UWGAN \cite{35-wang2019uwgan} &2.7023 &2.4759 &5.9846 &0.5463 &6.2201 &3.0632 &2.0544 &6.5840 &0.5012 &7.8833 &2.5412 &0.0988 &5.4089 &0.4965 &6.6929 \\
UIEC2Net \cite{36-wang2021uiec} &3.0829 &5.7149 &6.8495 &{\color[HTML]{FF0000} 0.6003} &{\color[HTML]{FF0000} 5.8255} & 3.2384 &4.1728 &6.9997 &0.5664 &9.0063 &3.0361 &2.7452 &6.3757 &0.5454 &6.0089 \\
NU2Net \cite{22-guo2023underwater} &{\color[HTML]{FF0000} 3.2404} &{\color[HTML]{0000FF} 5.8683} &{\color[HTML]{FF0000} 7.0905} &0.5940 &6.5328 &{\color[HTML]{0000FF}3.2519} &4.1084 &7.0111 &0.5515 &8.4507 & 3.0666 &2.6284 &6.3751 &0.5362 &8.0320 \\
U-shape \cite{17-peng2023u} &3.1043 &4.8119 &6.5588 &0.5673 &7.2762 &3.1100 &2.8325 &6.6533 &0.5201 &7.8776 &3.0072 &3.3378 &6.2638 &0.5487 &6.5036 \\
LSNet &3.0213 &4.0698  &{\color[HTML]{0000FF}7.0235} &0.5611  &8.2175 &{\color[HTML]{FF0000}3.2669} &2.7707 &{\color[HTML]{0000FF}7.0523} &0.5107 &10.3392 &{\color[HTML]{0000FF}3.1343} &3.2441 &{\color[HTML]{0000FF} 6.6566} &{\color[HTML]{0000FF} 0.5611}  &{\color[HTML]{FF0000} 5.6698}\\ \hline
\end{tabular}
}
\label{tab:table2}
\end{table}

\begin{figure}[H]
\centering
  \includegraphics[height=10cm]{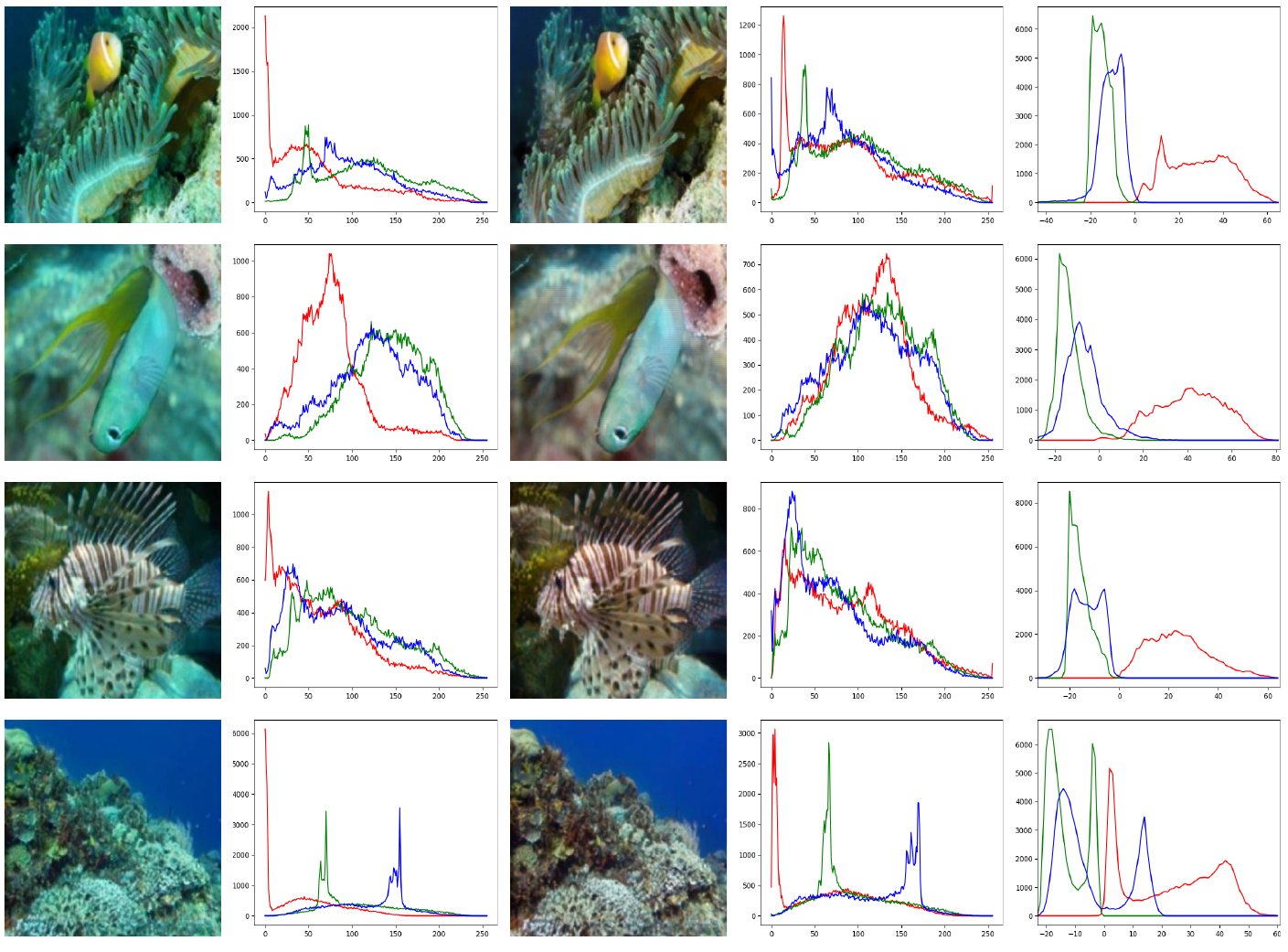}
  \caption{The visualization of the analysis and effectiveness of the LSNet. It can be seen that the red channel and some of the blue channel need to be compensated, while the green channel and most of the blue channel need to be suppressed.}
  \label{fig:mat}
\end{figure}
\subsection{Efficiency and Limitations Analysis of the LSNet}
The histogram of the original image of LSNet, the enhancement result, and the compensation image minus the over exposure attenuation image are visualized in Fig. \ref{fig:mat} to explore the reasons for the high efficiency of LSNet. 

It can be seen from Fig. \ref{fig:mat} that the values of the compensation image minus the over exposure attenuation image of the red channel of almost all images and the blue channel of some images are positive (The values of the red curves are concentrated in the part greater than 0 in the histogram of the fifth column). This confirms that in underwater images, the red channel has the largest attenuation compared to the green and blue channels. However, the values of the green and blue channels should be attenuated in most cases because they are overexposed (The values of the curves are concentrated in the part less than 0 in the histogram of the fifth column). This also matches some excellent traditional algorithms that use exposure images for underwater images. 

However, the LSNet still has its limitations, which are caused by too few parameters. Since LSNet divides the entire image into different regions, each region shares the same parameters. Therefore, in some scenarios, there will be discontinuities between region blocks of LSNet's output image, similar to CLAHE. This is caused by limited image regions. Therefore, when the number of dimensions in top-k increases, this problem may be solved. 

Secondly, unlike the interpretability of transmission map theory, this paper found that the areas of attention were not completely similar to transmission maps by the top k attention maps visualization. Transmission map theory is based on that scattering and absorption are related to distance, wavelength, and medium. However, this paper finds that the attention mechanism connects some areas of different colors and distances. This paper thinks that there may be three reasons for this: (1) The equation based on \ref{eq:eq1} has multiple unknown variables and only the original image is available. Therefore there will be multiple solutions. (2) Randomly initialized parameters make the model fall into a local optimal solution. (3) The imperfect reference image with color cast enables the model to learn the parameters that generate the color cast result. In future research, more restrictions on the equation \ref{eq:eq1} will be explored to obtain more interpretable and color-bias-free results.

\subsection{Ablation}

\begin{figure}[htbp]
  \begin{minipage}[t]{0.5\textwidth}
  \vspace{0.5cm} %
    \resizebox{\textwidth}{!}{%
\begin{tabular}{ccccc}
\hline
dataset                                          & model                         & UIQM   & UISM   & UICQE  \\ \hline
\multicolumn{1}{c|}{\multirow{5}{*}{Test-U}}     & \multicolumn{1}{c|}{LSNet}    & 3.0213 & 7.0235 & 0.5611 \\
\multicolumn{1}{c|}{}                            & \multicolumn{1}{c|}{wo\_x}    & 2.9749 & 6.8933 & 0.5517 \\
\multicolumn{1}{c|}{}                            & \multicolumn{1}{c|}{wo\_dx}   & 2.9984 & 6.7955 & 0.5480 \\
\multicolumn{1}{c|}{}                            & \multicolumn{1}{c|}{wo\_ox}   & 2.9924 & 6.7963 & 0.5458 \\
\multicolumn{1}{c|}{}                            & \multicolumn{1}{c|}{wo\_topk} & 2.8532 & 6.5154 & 0.5393 \\ \hline
\multicolumn{1}{c|}{\multirow{5}{*}{Test-blue}}  & \multicolumn{1}{c|}{LSNet}    & 3.2669 & 7.0523 & 0.5107 \\
\multicolumn{1}{c|}{}                            & \multicolumn{1}{c|}{wo\_x}    & 3.2259 & 7.0315 & 0.5034 \\
\multicolumn{1}{c|}{}                            & \multicolumn{1}{c|}{wo\_dx}   & 3.2179 & 6.8388 & 0.5099 \\
\multicolumn{1}{c|}{}                            & \multicolumn{1}{c|}{wo\_ox}   & 3.2393 & 6.9811 & 0.5025 \\
\multicolumn{1}{c|}{}                            & \multicolumn{1}{c|}{wo\_topk} & 3.0715 & 6.7494 & 0.4987 \\ \hline
\multicolumn{1}{c|}{\multirow{5}{*}{Test-green}} & \multicolumn{1}{c|}{LSNet}    & 3.1343 & 6.6566 & 0.5611 \\
\multicolumn{1}{c|}{}                            & \multicolumn{1}{c|}{wo\_x}    & 3.2183 & 6.4386 & 0.5586 \\
\multicolumn{1}{c|}{}                            & \multicolumn{1}{c|}{wo\_dx}   & 2.8533 & 5.4604 & 0.5599 \\
\multicolumn{1}{c|}{}                            & \multicolumn{1}{c|}{wo\_ox}   & 2.9840 & 6.0453 & 0.5106 \\
\multicolumn{1}{c|}{}                            & \multicolumn{1}{c|}{wo\_topk} & 2.7274 & 5.3250 & 0.4754 \\ \hline
\end{tabular}
}
\end{minipage}%
  \begin{minipage}[t]{0.65\textwidth}
    \vspace{-3.0cm} %
    \hspace{-0.3cm} %
    {\includegraphics[width=\textwidth]{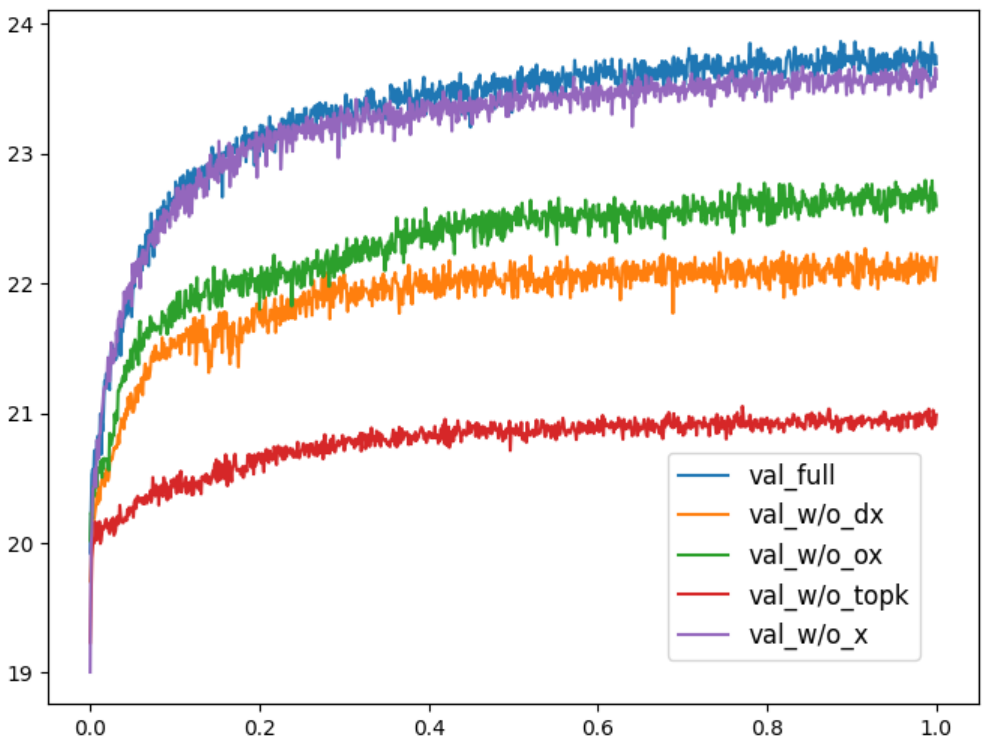}}
  \end{minipage}%
    \setlength{\abovecaptionskip}{-3.3cm}
    \setlength{\belowcaptionskip}{-0.3cm}
    \caption{The ablation results of the LSNet. The above figure is the validation PSNR results during the training process.}
\label{fig:ablation}
\end{figure}

To validate the effectiveness of the proposed module and the original image, compensation image, and over exposure attenuation image, the ablation experiments were conducted, as shown in Fig. \ref{fig:ablation}. In the table of Fig. \ref{fig:ablation}, the w/o\_ox represents the LSNet without the over exposure attenuation image, the w/o\_dx represents LSNet without the compensation image, the w/o\_x represents the LSNet without the original image, the w/o\_top\_k represents the LSNet without the top k module. In the figure of Fig. \ref{fig:ablation}, the val\_w/o\_ox represents the validation results of the LSNet during the training process without the over exposure attenuation image, the val\_w/o\_dx represents the validation results of the LSNet during the training process without the compensation image; the val\_w/o\_x the validation results of the LSNet during the training process without the original image, the val\_w/o\_top\_k represents the validation results of the LSNet during the training process without the top k module.
In Fig. \ref{fig:ablation}, it can be seen that all parts of the LSNet have an impact on the final results. For all of them, the top k module is the most important part because it is the core of this model. Secondly, the compensation image is the second important module, which also meets with the hypothesis of this paper. That is, most images underwater have attenuation, so the core of the underwater image enhancement task is compensation. Next, the third important part is the over exposure attenuation image. In most cases, the green and blue channels have an over-exposed problem, so the over exposure attenuation image is not as important as the compensation image. The use of the original image can also affect the results of the model, which shows that learning residual information is easier than learning the results directly.

\section{Conclusion}
\label{sec:con}
This paper proposes an extremely low-parameter underwater image enhancement model, named LSNet. This model uses a novel idea to enhance underwater images by generating the compensation image and the over exposure attenuation image of underwater images. It uses a top-k selection mechanism to find regions with similar attenuation and exposure to reduce parameter reuse. LSNet avoids the reconstruction module by image channel splitting, which reduces the number of parameters without losing most details. Experimental results show that the proposed LSNet model achieves comparable results to state-of-the-art models on public datasets. Ablation experiments show the effectiveness of each proposed part in the LSNet network.
\bibliographystyle{plain}
\bibliography{main}

\begin{thebibliography}{10}

\bibitem{4-ancuti2017color}
Codruta~O Ancuti, Cosmin Ancuti, Christophe De~Vleeschouwer, and Philippe
  Bekaert.
\newblock Color balance and fusion for underwater image enhancement.
\newblock {\em IEEE Transactions on image processing}, 27(1):379--393, 2017.

\bibitem{11-anwar2018deep}
Saeed Anwar, Chongyi Li, and Fatih Porikli.
\newblock Deep underwater image enhancement.
\newblock {\em arXiv preprint arXiv:1807.03528}, 2018.

\bibitem{8-drews2013transmission}
Paul Drews, Erickson Nascimento, Filipe Moraes, Silvia Botelho, and Mario
  Campos.
\newblock Transmission estimation in underwater single images.
\newblock In {\em Proceedings of the IEEE international conference on computer
  vision workshops}, pages 825--830, 2013.

\bibitem{16-fu2022uncertainty}
Zhenqi Fu, Wu~Wang, Yue Huang, Xinghao Ding, and Kai-Kuang Ma.
\newblock Uncertainty inspired underwater image enhancement.
\newblock In {\em European conference on computer vision}, pages 465--482.
  Springer, 2022.

\bibitem{22-guo2023underwater}
Chunle Guo, Ruiqi Wu, Xin Jin, Linghao Han, Weidong Zhang, Zhi Chai, and
  Chongyi Li.
\newblock Underwater ranker: Learn which is better and how to be better.
\newblock In {\em Proceedings of the AAAI conference on artificial
  intelligence}, volume~37, pages 702--709, 2023.

\bibitem{13-guo2019underwater}
Yecai Guo, Hanyu Li, and Peixian Zhuang.
\newblock Underwater image enhancement using a multiscale dense generative
  adversarial network.
\newblock {\em IEEE Journal of Oceanic Engineering}, 45(3):862--870, 2019.

\bibitem{30-he2010single}
Kaiming He, Jian Sun, and Xiaoou Tang.
\newblock Single image haze removal using dark channel prior.
\newblock {\em IEEE transactions on pattern analysis and machine intelligence},
  33(12):2341--2353, 2010.

\bibitem{3-hitam2013mixture}
Muhammad~Suzuri Hitam, Ezmahamrul~Afreen Awalludin, Wan Nural Jawahir Hj~Wan
  Yussof, and Zainuddin Bachok.
\newblock Mixture contrast limited adaptive histogram equalization for
  underwater image enhancement.
\newblock In {\em 2013 International conference on computer applications
  technology (ICCAT)}, pages 1--5. IEEE, 2013.

\bibitem{14-islam2020fast}
Md~Jahidul Islam, Youya Xia, and Junaed Sattar.
\newblock Fast underwater image enhancement for improved visual perception.
\newblock {\em IEEE Robotics and Automation Letters}, 5(2):3227--3234, 2020.

\bibitem{37-jaffe1990computer}
Jules~S Jaffe.
\newblock Computer modeling and the design of optimal underwater imaging
  systems.
\newblock {\em IEEE Journal of Oceanic Engineering}, 15(2):101--111, 1990.

\bibitem{1-jian2021underwater}
Muwei Jian, Xiangyu Liu, Hanjiang Luo, Xiangwei Lu, Hui Yu, and Junyu Dong.
\newblock Underwater image processing and analysis: A review.
\newblock {\em Signal Processing: Image Communication}, 91:116088, 2021.

\bibitem{23-jiang2023five}
Jingxia Jiang, Tian Ye, Jinbin Bai, Sixiang Chen, Wenhao Chai, Shi Jun, Yun
  Liu, and Erkang Chen.
\newblock Five a+ network: You only need 9k parameters for underwater image
  enhancement.
\newblock {\em arXiv preprint arXiv:2305.08824}, 2023.

\bibitem{15-li2021underwater}
Chongyi Li, Saeed Anwar, Junhui Hou, Runmin Cong, Chunle Guo, and Wenqi Ren.
\newblock Underwater image enhancement via medium transmission-guided
  multi-color space embedding.
\newblock {\em IEEE Transactions on Image Processing}, 30:4985--5000, 2021.

\bibitem{34-li2020underwater}
Chongyi Li, Saeed Anwar, and Fatih Porikli.
\newblock Underwater scene prior inspired deep underwater image and video
  enhancement.
\newblock {\em Pattern Recognition}, 98:107038, 2020.

\bibitem{12-li2019underwater}
Chongyi Li, Chunle Guo, Wenqi Ren, Runmin Cong, Junhui Hou, Sam Kwong, and
  Dacheng Tao.
\newblock An underwater image enhancement benchmark dataset and beyond.
\newblock {\em IEEE transactions on image processing}, 29:4376--4389, 2019.

\bibitem{5-li2015underwater}
Chongyi Li and Jichang Guo.
\newblock Underwater image enhancement by dehazing and color correction.
\newblock {\em Journal of Electronic Imaging}, 24(3):033023--033023, 2015.

\bibitem{41-li2019fusion}
Hanyu Li, Jingjing Li, and Wei Wang.
\newblock A fusion adversarial underwater image enhancement network with a
  public test dataset.
\newblock {\em arXiv preprint arXiv:1906.06819}, 2019.

\bibitem{40-liu2020real}
Risheng Liu, Xin Fan, Ming Zhu, Minjun Hou, and Zhongxuan Luo.
\newblock Real-world underwater enhancement: Challenges, benchmarks, and
  solutions under natural light.
\newblock {\em IEEE transactions on circuits and systems for video technology},
  30(12):4861--4875, 2020.

\bibitem{19-lyu2022efficient}
Zhangkai Lyu, Andrew Peng, Qingwei Wang, and Dandan Ding.
\newblock An efficient learning-based method for underwater image enhancement.
\newblock {\em Displays}, 74:102174, 2022.

\bibitem{38-mcglamery1980computer}
BL~McGlamery.
\newblock A computer model for underwater camera systems.
\newblock In {\em Ocean Optics VI}, volume 208, pages 221--231. SPIE, 1980.

\bibitem{25-mittal2012making}
Anish Mittal, Rajiv Soundararajan, and Alan~C Bovik.
\newblock Making a “completely blind” image quality analyzer.
\newblock {\em IEEE Signal processing letters}, 20(3):209--212, 2012.

\bibitem{20-naik2021shallow}
Ankita Naik, Apurva Swarnakar, and Kartik Mittal.
\newblock Shallow-uwnet: Compressed model for underwater image enhancement
  (student abstract).
\newblock In {\em Proceedings of the AAAI Conference on Artificial
  Intelligence}, volume~35, pages 15853--15854, 2021.

\bibitem{28-panetta2015human}
Karen Panetta, Chen Gao, and Sos Agaian.
\newblock Human-visual-system-inspired underwater image quality measures.
\newblock {\em IEEE Journal of Oceanic Engineering}, 41(3):541--551, 2015.

\bibitem{17-peng2023u}
Lintao Peng, Chunli Zhu, and Liheng Bian.
\newblock U-shape transformer for underwater image enhancement.
\newblock {\em IEEE Transactions on Image Processing}, 2023.

\bibitem{9-perez2017deep}
Javier Perez, Aleks~C Attanasio, Nataliya Nechyporenko, and Pedro~J Sanz.
\newblock A deep learning approach for underwater image enhancement.
\newblock In {\em Biomedical Applications Based on Natural and Artificial
  Computing: International Work-Conference on the Interplay Between Natural and
  Artificial Computation, IWINAC 2017, Corunna, Spain, June 19-23, 2017,
  Proceedings, Part II}, pages 183--192. Springer, 2017.

\bibitem{31-rao2023deep}
Yuan Rao, Wenjie Liu, Kunqian Li, Hao Fan, Sen Wang, and Junyu Dong.
\newblock Deep color compensation for generalized underwater image enhancement.
\newblock {\em IEEE Transactions on Circuits and Systems for Video Technology},
  2023.

\bibitem{6-sathya2015underwater}
R~Sathya, M~Bharathi, and G~Dhivyasri.
\newblock Underwater image enhancement by dark channel prior.
\newblock In {\em 2015 2nd International Conference on Electronics and
  Communication Systems (ICECS)}, pages 1119--1123. IEEE, 2015.

\bibitem{39-schechner2005recovery}
Yoav~Y Schechner and Nir Karpel.
\newblock Recovery of underwater visibility and structure by polarization
  analysis.
\newblock {\em IEEE Journal of oceanic engineering}, 30(3):570--587, 2005.

\bibitem{2-shi2022integrating}
Zhenghao Shi, Yongli Wang, Zhaorun Zhou, and Wenqi Ren.
\newblock Integrating deep learning and traditional image enhancement
  techniques for underwater image enhancement.
\newblock {\em IET Image Processing}, 16(13):3471--3484, 2022.

\bibitem{35-wang2019uwgan}
Nan Wang, Yabin Zhou, Fenglei Han, Haitao Zhu, and Jingzheng Yao.
\newblock Uwgan: underwater gan for real-world underwater color restoration and
  dehazing.
\newblock {\em arXiv preprint arXiv:1912.10269}, 2019.

\bibitem{29-wang2019experimental}
Yan Wang, Wei Song, Giancarlo Fortino, Li-Zhe Qi, Wenqiang Zhang, and Antonio
  Liotta.
\newblock An experimental-based review of image enhancement and image
  restoration methods for underwater imaging.
\newblock {\em IEEE Access}, 7:140233--140251, 2019.

\bibitem{10-wang2017deep}
Yang Wang, Jing Zhang, Yang Cao, and Zengfu Wang.
\newblock A deep cnn method for underwater image enhancement.
\newblock In {\em 2017 IEEE international conference on image processing
  (ICIP)}, pages 1382--1386. IEEE, 2017.

\bibitem{36-wang2021uiec}
Yudong Wang, Jichang Guo, Huan Gao, and Huihui Yue.
\newblock Uiec\^{} 2-net: Cnn-based underwater image enhancement using two
  color space.
\newblock {\em Signal Processing: Image Communication}, 96:116250, 2021.

\bibitem{26-wang2004image}
Zhou Wang, Alan~C Bovik, Hamid~R Sheikh, and Eero~P Simoncelli.
\newblock Image quality assessment: from error visibility to structural
  similarity.
\newblock {\em IEEE transactions on image processing}, 13(4):600--612, 2004.

\bibitem{7-wen2013single}
Haocheng Wen, Yonghong Tian, Tiejun Huang, and Wen Gao.
\newblock Single underwater image enhancement with a new optical model.
\newblock In {\em 2013 IEEE International Symposium on Circuits and Systems
  (ISCAS)}, pages 753--756. IEEE, 2013.

\bibitem{18-yang2021laffnet}
Hao-Hsiang Yang, Kuan-Chih Huang, and Wei-Ting Chen.
\newblock Laffnet: A lightweight adaptive feature fusion network for underwater
  image enhancement.
\newblock In {\em 2021 IEEE international conference on robotics and automation
  (ICRA)}, pages 685--692. IEEE, 2021.

\bibitem{27-yang2015underwater}
Miao Yang and Arcot Sowmya.
\newblock An underwater color image quality evaluation metric.
\newblock {\em IEEE Transactions on Image Processing}, 24(12):6062--6071, 2015.

\bibitem{24-zhu2023biformer}
Lei Zhu, Xinjiang Wang, Zhanghan Ke, Wayne Zhang, and Rynson~WH Lau.
\newblock Biformer: Vision transformer with bi-level routing attention.
\newblock In {\em Proceedings of the IEEE/CVF conference on computer vision and
  pattern recognition}, pages 10323--10333, 2023.

\end{thebibliography}
\end{document}